\documentclass[11pt]{article}
\usepackage[margin=0.9in]{geometry}
\usepackage{amsmath,booktabs,graphicx,url}
\makeatletter
\newcommand{\centerfloatcaption}{%
  \long\def\@makecaption##1##2{%
    \vskip\abovecaptionskip
    {\centering ##1: ##2\par}%
    \vskip\belowcaptionskip}}
\makeatother
\newcommand{\T}[2]{{}^{#1}\!T_{#2}}
\title{Robot View: Calibrated RGB-D and LiDAR Mapping Across Viewer Restarts on an Articulated Mobile Robot}
\author{Junyao He\\{\small Independent Researcher} \and Zaide Zhang}
\date{}

\begin{document}
\maketitle
\begin{abstract}
Inspection robots commonly mount their RGB-D cameras on an articulated head while the LiDAR remains fixed to the chassis, so a single rigid extrinsic between the two sensors is valid only at the joint configuration in which it was measured. In this paper, we present Robot View, a mapping and inspection system for the XR2 wheeled humanoid that maintains a persistent RGB-D and LiDAR map across viewer restarts. The camera--LiDAR mount is estimated at a reference posture from a planar checkerboard with a known outer boundary, using residuals on both the board plane and its measured outline; the system then uses forward kinematics to apply this correction at other head postures. Surface appearance is retained as oriented Gaussian primitives, range measurements are kept in a short-lived scan buffer alongside a persistent voxel archive, and a reopened map remains read-only until two independently timestamped scans agree with the archived geometry. On the six fixed-posture board placements used to select the correction, the mean per-placement 95th percentile of absolute point-to-plane error decreases from 206.52 to 23.53\,mm, and five of the six placements satisfy all three acceptance criteria. The calibration is evaluated quantitatively, while the mapping and recovery components are demonstrated on the robot; their quantitative evaluation is left to future work.
\end{abstract}

\section{Introduction}

A robot that inspects a building passes the same doorway many times, and each pass leaves more than one record of it: a camera captures what the jamb looks like, a LiDAR measures where the threshold is. The records differ in range, resolution and rate, and neither is complete alone. Combining them turns a routine question---whether this is the doorway seen last week---into a geometric one: by how much must a new observation be displaced to agree with the stored one, and how much disagreement is acceptable? This paper concerns a mapping system that answers it on a robot whose camera and LiDAR share no fixed relationship.

Robot View runs on the XR2, a wheeled humanoid whose torso carries an articulated head. The RGB-D camera turns with the head; the LiDAR stays on the chassis. The arrangement is a choice rather than a constraint: a fixed sensor sees only what the chassis sees, whereas the head lets the camera look into shelves, under tables and above the plane the LiDAR sweeps. Its price is the calibration problem addressed here, and immobilising the head during mapping would give up exactly the views that make the map worth keeping. A rigid transform between the two sensors is therefore valid only at the configuration in which it was measured, and since forward kinematics propagates the camera pose, a mounting error at the reference posture is repeated at every posture the head later adopts. The robot already estimates its own motion, and the system builds on those estimates rather than replacing them. The map inherits the accuracy of the platform's state estimation, and the one relationship the platform does not estimate---the camera mount---becomes the quantity this paper has to measure.

For an inspection robot, the practical value of a stored map is that it permits comparison over time. If the archive records a closed door and a current scan records an open one, the operator has a finding; if the archive has drifted, the same comparison yields an artefact, and the display offers no way to distinguish the two. Two properties follow that a conventional map does not provide. An archived surface must stay identifiable as archive rather than as current evidence, so that the operator knows what the robot has just measured. And a current scan must be placed correctly before it is allowed to become archive, because a registration error admitted once is afterwards indistinguishable from something observed.

A harder requirement appears when the viewer is stopped and reopened. A stored map outlives the process that built it, but that process leaves state which cannot be restored exactly: the optimizer restarts empty, the scan buffer refills, and the pose at reopen is at best an approximation of the pose at shutdown. A heading error of a few degrees is enough to place a new wall beside the archived one, and once such a scan is integrated it is hard to separate from real structure. Reopening is therefore not the continuation of a session but a fresh dialogue with an archive whose frame must be re-established; the archive stays read-only until the current observations justify otherwise.

Three decisions follow, and each is a contribution. The camera mount is estimated at a reference posture against a physical target and carried through the kinematic chain; the residual combines the target's plane with its outer outline, so displacement through the board can be separated from displacement along it, and nominal and corrected mounts are compared on identical returns, which makes the improvement directly attributable. Appearance and range sit in separate layers with different lifetimes, so a surface keeps its colour after the camera turns away while a scan that is no longer current stops being presented as a measurement. Admission to a reopened map requires two independently timestamped alignments to agree, which addresses the failure that matters in corridors whose walls repeat. The calibration is evaluated in Section~\ref{sec:exp}; the other two components run on the robot and are presented through their design, and Section~\ref{sec:disc} states what evidence would quantify them.

Section~\ref{sec:rel} reviews related work, Section~\ref{sec:sys} fixes the platform and the alignments in time and in frame, Section~\ref{sec:method} presents the three components, and Section~\ref{sec:exp} reports the calibration experiment. Section~\ref{sec:doorway} compares the two map representations at an office doorway, before Section~\ref{sec:disc} discusses the results and limitations.

\section{Related work}\label{sec:rel}

Three literatures bear on this problem: how dense maps are represented, how a camera is registered to a LiDAR on a platform whose geometry changes, and how a map is reused after its session ends. In each case the present design departs from the work at a point that follows from one fact---here the poses already exist.

\subsection{Representing and persisting a dense map}

Dense reconstruction systems differ in what they preserve when the sensor moves on. BundleFusion re-integrates the surface whenever global pose estimates change~\cite{bundle}; Voxblox derives a signed distance field for planning from an integrated surface~\cite{voxblox}, and nvblox performs the same incremental update on the GPU~\cite{nvblox}. These representations answer geometric questions well and appearance questions poorly.

Gaussian splatting changes that balance: a scene is held as primitives with position, anisotropic extent, orientation and colour, and rendered by projecting them~\cite{kerbl}. SplaTAM couples such a map to camera tracking~\cite{splatam}, and Matsuki et al.\ do the same for monocular SLAM~\cite{monogs}. Pose is estimated jointly with the map in all of these systems. On the XR2 the poses are supplied, so the archive need not support tracking; it must instead retain enough per-primitive structure to draw a surface again as it was seen, and keep that structure distinguishable from what the current session has built. The representation in Section~\ref{sec:layers} is organised around that distinction.

\subsection{Registering a camera to a LiDAR on a platform that moves}

Multi-sensor systems that estimate motion from both modalities assume a rig of fixed geometry. FAST-LIVO2 fuses inertial, LiDAR and image measurements in one estimator~\cite{fastlivo}, and GS-LIVO pairs multisensor odometry with Gaussian mapping~\cite{gslivo}. External calibration is usually handled separately: Zhou et al.\ establish camera--LiDAR correspondences from lines and planes on a target~\cite{zhou}, and Koide et al.\ obtain an extrinsic from a single targetless observation~\cite{koide}.

The XR2 differs in that the camera moves relative to the LiDAR during operation, so the correction must remain valid as it moves. We estimate the mount at one posture and propagate it through forward kinematics, which makes the accuracy of the kinematic chain part of the calibration problem. The residual also differs from a plane-only formulation: a board facing the camera head-on constrains displacement along its normal far more strongly than translation within its plane, so a plane-only fit can slide without penalty. The board's measured outline supplies the missing in-plane constraint, in the spirit of~\cite{zhou} but from a single passive target of known extent.

\subsection{Reusing a map across sessions}

Reusing a map requires deciding where the current session falls within the earlier one. Scan Context retrieves candidate places from a scan~\cite{scancontext}, and ORB-SLAM3 merges maps so that correspondences can be established between them~\cite{orb}. KISS-ICP aligns a new scan to a nearby portion of a map and shows that point-to-point registration suffices when the initial guess is close~\cite{kiss}.

Robot View addresses the case these methods leave to the operator. The archive is already chosen and an approximate pose is available, so what remains is a local, binary decision about whether the current observations support writing at this location. The cost is asymmetric: refusing wastes time, while writing incorrectly corrupts the map for every later session. The rule in Section~\ref{sec:recover} therefore targets the second error and accepts a longer delay.

\section{System overview}\label{sec:sys}

Every observation must satisfy two conditions before it can enter a map: it must be bound to the instant at which it was taken, and expressed in a frame the map understands. This section fixes the platform, the timing rules, and the composition that turns a joint configuration and a chassis pose into a camera pose.

\subsection{Platform and data flow}

\begin{figure}[t]
\centering
\includegraphics[width=\linewidth]{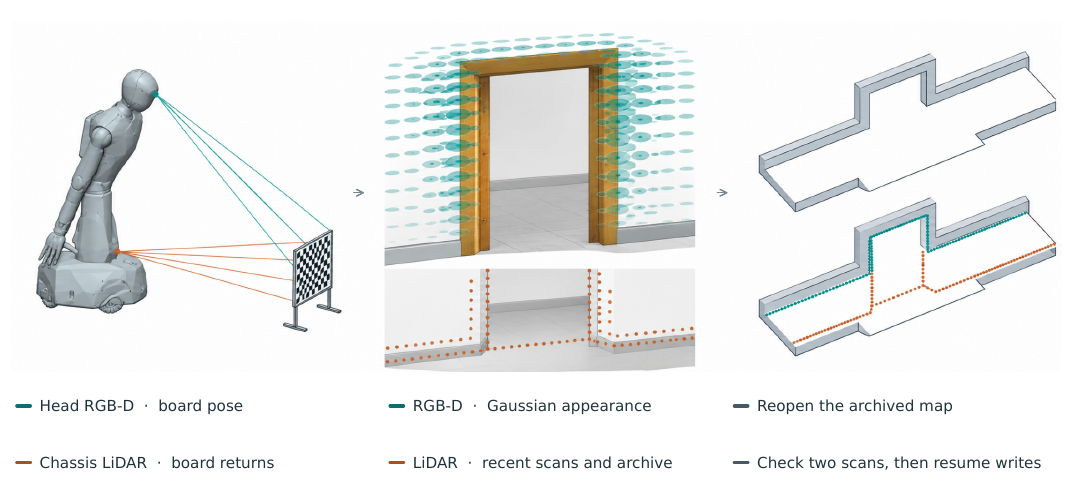}\par\vspace{2pt}
\begin{minipage}[t]{.329\linewidth}\centering\small (a) Sensor calibration\end{minipage}\hfill\begin{minipage}[t]{.329\linewidth}\centering\small (b) Layered mapping\end{minipage}\hfill\begin{minipage}[t]{.329\linewidth}\centering\small (c) Map reuse after restart\end{minipage}
\caption{Robot View from acquisition to map reuse. (a) The head RGB-D camera and chassis LiDAR observe a shared target to estimate the mount correction at a reference posture. (b) RGB-D observations supply Gaussian appearance, while scans supply recent geometry and a persistent reference. (c) The archived map is visible after restart. Writing resumes once two independently timestamped scan alignments agree. The panels are system schematics drawn with assistance from ImageGen.}\label{fig:architecture}
\end{figure}

The system runs on the onboard Thor computer over two ROS~2 domains, which keep the robot's publishers separate from the chassis stack: the body domain supplies colour, depth, intrinsics, visual odometry and joint feedback, the chassis domain wheel odometry, navigation transforms, paths and Livox scans. Robot View subscribes to both without republishing the robot's transforms, serves the assembled scene over local HTTP to a WebGL viewer, and leaves planning to the existing navigation stack. Figure~\ref{fig:architecture}(a) shows how a common target ties the sensors together. The corrected observations enter separate map layers in Fig.~\ref{fig:architecture}(b), which provide the reference for the restart check in Fig.~\ref{fig:architecture}(c). Since the system adds no sensor and no estimator, any improvement in the map comes from how existing measurements are related, and any error originating upstream persists.

\subsection{Temporal alignment}

The first source of misalignment is motion between measurements. The system treats the depth frame as the reference: an alignment node pairs it with the colour image whose timestamp is closest within 50\,ms, reprojects colour into the depth camera, masks the robot's own geometry, and republishes the pair under one stamp. Visual odometry and chassis transforms are then looked up at that stamp, while buffered wheel odometry and scans use the nearest sample within 140\,ms. Joint angles are taken from the most recent feedback, because the controller publishes them as state rather than as a trajectory. The consequence is an error that grows with head velocity: during a fast turn, an image and its visual odometry can agree in time while the angles used to compute the mount are already stale. The cuVSLAM and nvblox branches run alongside and supply visual odometry and a diagnostic mesh.

\subsection{Pose composition across frames}

A shared timestamp still leaves a choice of spatial origin. Write $\T{A}{B}$ for the transform taking coordinates from frame $B$ to frame $A$. The chassis pose is published in both the wheel odometry frame and the navigation frame, and the two need not share an origin; if $\T{O}{B}$ and $\T{N}{B}$ describe the same base at a matched instant, the viewer estimates $\T{N}{O}=\T{N}{B}(\T{O}{B})^{-1}$ and can express a fallback pose in the frame the map was established in. With chassis data available, the world camera pose is the base pose composed with the corrected mount of Section~\ref{sec:calib}; simultaneous visual odometry anchors the visual frame to that pose and substitutes when chassis input is missing.

\section{Method}\label{sec:method}

The three components address the placement of an observation at three stages, and each relies on a different kind of evidence. Calibration fixes the relation between the sensors from a target seen at a known posture. The layered map decides what is retained, and for how long, once an observation has a pose. Admission decides whether a pose estimated after a restart is trustworthy enough to write with. They are kept separate on purpose: a mount error, a storage error and an admission error all reach the operator as a misaligned or misleading surface in the same rendered image, and only the first is repaired by calibration.

\subsection{Calibrating the head camera against the chassis LiDAR}\label{sec:calib}

Forward kinematics describes how the camera moves with the head, not how it is attached, and that distinction is why a calibration is needed. A mounting error at the reference posture enters every frame the camera contributes, and because the same correction is propagated to other postures, motion does not average it away: it displaces the whole visual layer relative to the range layer, uniformly and persistently. Two sensors that disagree by several centimetres make a doorframe appear doubled, and the doubling looks like a timing fault until the timestamps have been checked.

The target is a metal-backed checkerboard with a $400\times300$\,mm outer boundary, twelve by nine squares of 30\,mm pitch and $11\times8$ internal corners. A board with both a printed pattern and a physical edge is chosen deliberately, because the two sensors see different features of the same object. The camera detects corners in the colour image and estimates the board pose by perspective-$n$-point fitting with the calibrated intrinsics, then refines it against measured depth. The LiDAR contributes returns from the physical face, accumulated over several revolutions. Since the printed area is smaller than the backing, treating the print boundary as the physical target would predict the wrong extent for those returns and absorb the resulting bias silently into the extrinsic.

\begin{figure}[t]\centering
\includegraphics[width=\linewidth]{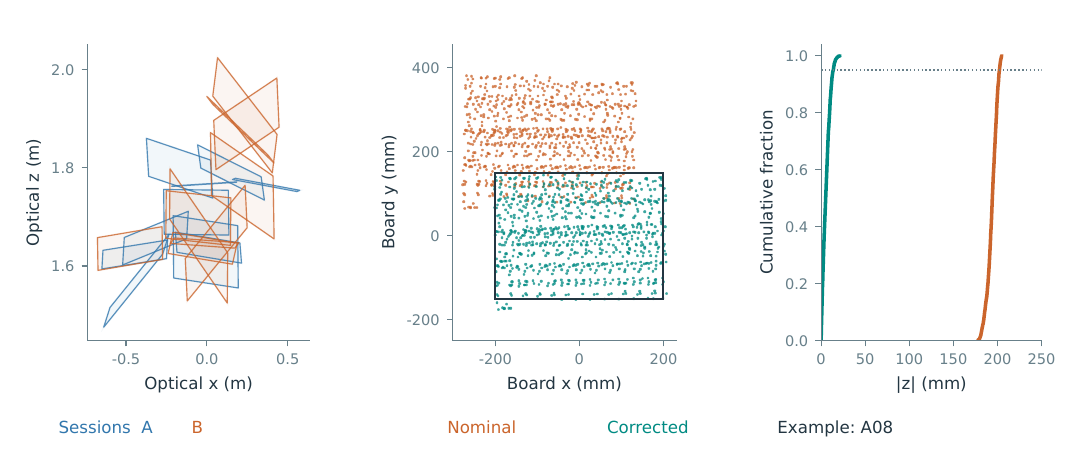}\par\vspace{2pt}
\begin{minipage}[t]{.329\linewidth}\centering\small (a) Measured board poses\end{minipage}\hfill\begin{minipage}[t]{.329\linewidth}\centering\small (b) Returns in board coordinates\end{minipage}\hfill\begin{minipage}[t]{.329\linewidth}\centering\small (c) Distance to the board plane\end{minipage}
\caption{Calibration geometry from the two recording sessions. (a) All 20 board outlines projected onto the optical $xz$ plane. (b) LiDAR returns at A08 transformed into board coordinates with the nominal and corrected mounts; the rectangle marks the physical board edge. (c) Empirical distributions of absolute plane distance for the same returns, with the 95th percentile marked.}\label{fig:calibration}
\end{figure}

Capture proceeds with the head, waist and lift at a reference setting while the board is moved through ten placements per session. Each sample records the board pose, the returns in LiDAR coordinates, the nominal extrinsic and the joint configuration, so any later fit can be recomputed and any error traced to a placement. The first seven placements constrain the fit and the remaining three are held back, so that the fit can be inspected on poses it was not allowed to see (Fig.~\ref{fig:calibration}(a)). For placement A08, Fig.~\ref{fig:calibration}(b) shows the in-plane shift in the same returns, and Fig.~\ref{fig:calibration}(c) shows their distance from the board plane.

Residuals are constructed in the board frame, where the two kinds of error separate. Let $B_i=\T{C}{P_i}$ map board $i$ to the depth optical frame, $p_{ij}$ a homogeneous LiDAR point, and $X=\T{C}{L}$ the unknown transform at the reference posture. Then
\begin{equation}
 u_{ij}=B_i^{-1}X p_{ij}, \qquad
 m_{ik}=\tfrac12\big[Q_{.02}(\{u_{ij,k}\}_j)+Q_{.98}(\{u_{ij,k}\}_j)\big],\quad k\in\{x,y\}.
\end{equation}
The plane residual is the signed $z$ coordinate. The two components of $m_i$ estimate the midpoint of the observed outline along the board axes, using the 2nd and 98th percentiles rather than the extremes so that a few stray returns at an edge do not set the bound. This second residual keeps the fit from sliding: a board facing the camera head-on constrains translation along its normal far more strongly than within its plane, so a plane-only fit can trade in-plane displacement against rotation and still explain the data, whereas the outline resolves the trade. A partially observed edge will bias the midpoint even when most returns lie close to the plane, which is why centre error is reported separately from plane distance.

The implementation left-multiplies the nominal transform by a correction parameterised as a rotation vector and a translation. Per placement it selects 64 evenly indexed plane samples and repeats the two midpoint residuals 16 times, weighting the fit towards the outline. A soft-$L_1$ fit with a 12\,mm residual scale downweights returns that do not belong to the board face. A solution is rejected if the optimisation fails or if the ratio of smallest to largest singular value of its numerical Jacobian falls below 0.012, which happens when the footprint is too small or too oblique to constrain the transform. The fitted transform therefore depends on the observed footprint as much as on the plane normals.

The fitted transform describes a stationary head. To use it during motion we select a correction at the reference posture and carry it with the nominal mount. Two development sessions yielded two candidates, selected on six held-out placements against limits of 30\,mm on plane P95, 38\,mm on midpoint error and $4^\circ$ on normal angle. For nominal $X_0$ and selected $\hat X$ the runtime rule is
\begin{equation}
 \T{B}{C}(q)=\T{B}{C,\mathrm{nom}}(q)\,X_0\hat X^{-1}.
 \label{eq:mount}
\end{equation}
The viewer verifies the robot model, base link, camera link and optical frame before applying it, and because the correction is local, the transforms the robot publishes keep their owners. This encodes an assumption: that the kinematic chain carries a constant correction accurately to other configurations, which is plausible for a rigid head on a well-modelled neck and less so where the chain contains flex, joint offsets or feedback latency. It is what a single-posture experiment cannot test, and we return to it in Section~\ref{sec:disc}.

\subsection{A layered map: appearance, recent range and archive}\label{sec:layers}

Once the mount is corrected, both sensors place their observations of a doorframe at the same location, yet the observations differ in purpose and deserve different lifetimes. Colour remains useful after the camera has looked away, whereas a scan is useful only while it describes what the LiDAR has just measured. No single layer satisfies both, so the map is split into appearance as Gaussian primitives, a short-lived scan buffer, and a persistent voxel archive that provides the reference used after a restart.

Each valid depth sample is back-projected with its colour into the display frame. A CUDA fitter projects visible Gaussians into image tiles, sorts them by depth and adjusts colour, projected centre, tangential scale and opacity against pixel residuals, revisiting overlapping views from a window of up to eight keyframes. The model uses view-independent colour, which suits the surfaces this system inspects and keeps the per-primitive record compact. Where background depth has been observed, a depth comparison masks a transient foreground, so a person crossing the scene does not permanently alter the wall.

\begin{figure}[t]\centering
\includegraphics[width=\linewidth]{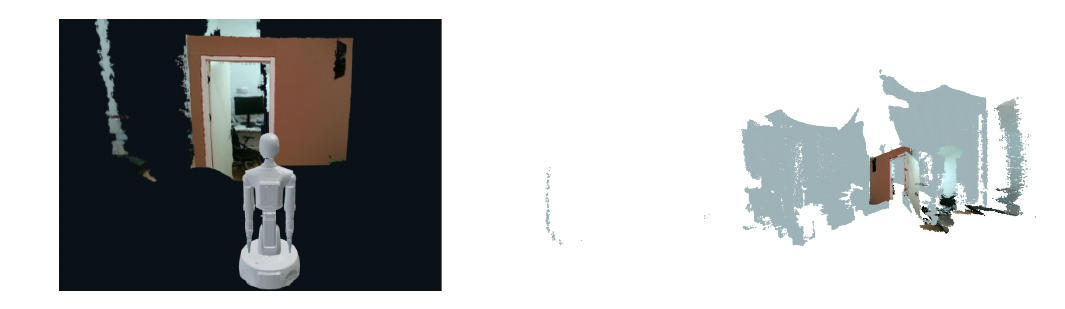}\par\vspace{2pt}
\begin{minipage}[t]{.49\linewidth}\centering\small (a) Appearance in the viewer\end{minipage}\hfill\begin{minipage}[t]{.49\linewidth}\centering\small (b) Scene point cloud\end{minipage}
\par\vspace{6pt}
\includegraphics[width=\linewidth]{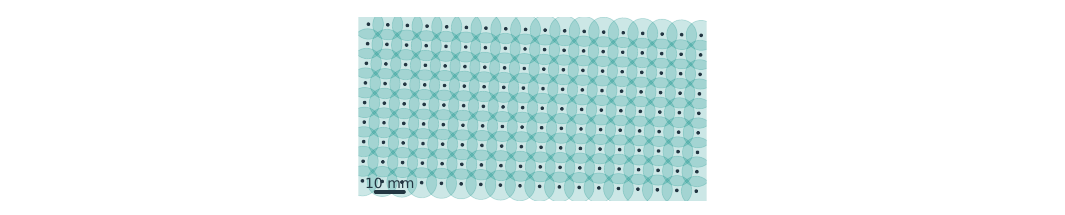}\par\vspace{2pt}
\begin{minipage}[t]{1\linewidth}\centering\small (c) Local Gaussian footprints in the archive\end{minipage}
\caption{Appearance and geometry in three development records. (a) Onboard Gaussian reconstruction with the viewer's robot overlay. (b) Oblique projection of a scene export, where points carrying photographic colour keep their RGB and the remainder are grey. (c) A small region of a native Gaussian snapshot projected in the $yz$ plane, with dots marking centres and ellipses showing tangent footprints from the stored scales and normals. Panels come from separate records and viewpoints.}\label{fig:maplayers}
\end{figure}

Figure~\ref{fig:maplayers}(a) shows a doorway rendered in the viewer. The oblique point-cloud view in Fig.~\ref{fig:maplayers}(b) shows the spatial coverage of a separate scene export. A coloured point says where a colour appears, but the extent and orientation of its footprint decide how much of the wall it covers from another direction. A native snapshot therefore stores, per primitive, centre, scalar scale, quantised normal, RGB and opacity (Fig.~\ref{fig:maplayers}(c)). On reload the renderer gives active primitives priority over historical ones within the same 2.5\,cm cell, and the operator can display the active model alone to see what the current run has contributed.

The range layer records what the robot has recently measured. Overlaying consecutive scans completes edges a single sweep only partly observed, but keeping them indefinitely would leave returns on the display after the robot has moved away, and the operator could not separate current geometry from remembered geometry. The buffer retains up to 32 scans over two seconds and drops those more than 25\,cm or $15^\circ$ from the current view; a gap of 1.25\,s expires the overlay. Independently, a voxel archive at 4\,cm stores positions and intensities for later inspection and for the registration of Section~\ref{sec:recover}. Isolated returns may vanish from the rendered view because neighbours are joined into surfaces; free-space decisions stay with the navigation stack. Table~\ref{tab:layers} lists what each of the four views draws.

\begin{table}[t]\centering\small
\centerfloatcaption
\caption{Four views of the same robot frame. The scan window is rebuilt from fresh returns after a restart.}\label{tab:layers}
\begin{tabular}{@{}ccc@{}}\toprule
View & Appearance & Range geometry\\\midrule
Combined scene & Active and archived Gaussians & Recent scans\\
Gaussian map & Active and archived Gaussians & None\\
Online reconstruction & Active fitter only & None\\
LiDAR map & None & Accumulated returns\\\bottomrule
\end{tabular}
\end{table}

\subsection{Admitting new observations to a reopened map}\label{sec:recover}

\begin{figure}[t]\centering
\centerfloatcaption
\includegraphics[width=\linewidth]{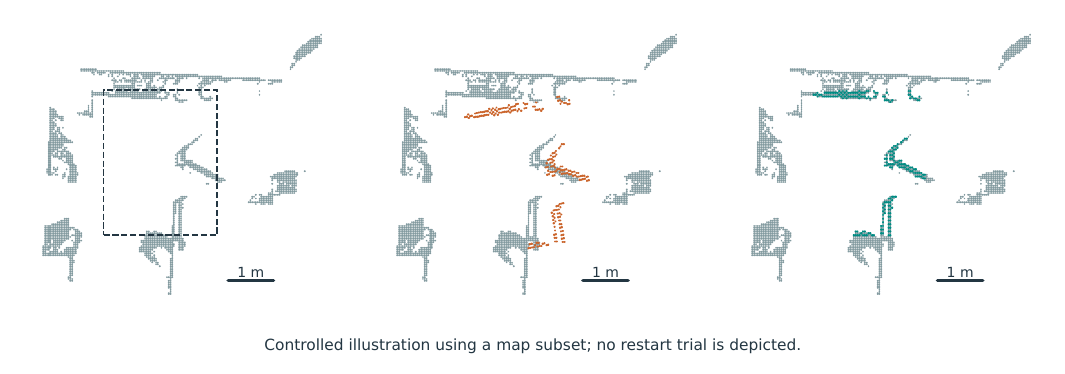}\par\vspace{2pt}
\begin{minipage}[t]{.329\linewidth}\centering\small (a) Archived geometry\end{minipage}\hfill\begin{minipage}[t]{.329\linewidth}\centering\small (b) Perturbed subset\end{minipage}\hfill\begin{minipage}[t]{.329\linewidth}\centering\small (c) After the inverse transform\end{minipage}
\caption{Geometry of a map-frame offset. (a) Plan view of scene points between 0.4 and 1.8\,m in height, with the dashed box selecting a subset. (b) The subset rotated $10^\circ$ about its centroid and translated by $(+0.32,-0.22)$\,m. (c) The known inverse restores its position. All three panels use archived points in a controlled illustration of a known offset.}\label{fig:recovery}
\end{figure}

Saving primitives and returns makes an old room visible as soon as its archive is opened, but a visible archive is not yet one that can be extended. Extending the room requires the one quantity the archive cannot contain: the pose of the first new scan in the archived frame. Figure~\ref{fig:recovery}(a) selects a subset of the archive. Applying a known offset in Fig.~\ref{fig:recovery}(b) and its inverse in Fig.~\ref{fig:recovery}(c) illustrates the frame change. The server keeps the map identifier, the last base pose, any navigation-frame alignment, the snapshots and the LiDAR blocks; closing the browser leaves this running, while restarting the service starts a fresh optimizer and exposes the previous snapshot as history.

The initial estimate comes from the best available source. Where the LiDAR archive holds enough geometry the matcher aligns the new scan to archived returns; otherwise archived Gaussian centres serve as the reference. A stored navigation alignment with the current navigation pose gives the seed when both exist, and the saved base pose is the fallback. The three differ in what they assume, and a failure in one is not evidence of a failure in another.

Matching follows local point-cloud registration. As in KISS-ICP~\cite{kiss}, nearby correspondences constrain the update; our matcher trims distant pairs, estimates translation with rotation about the vertical axis, and checks the number of supporting pairs, the median residual and the spatial spread of the support. The vertical-axis restriction reflects a platform that drives on a floor; the spread check rejects alignments supported by a single wall, which can be consistent and still leave position along that wall undetermined.

A single accepted alignment still leaves a plausible match to the wrong wall, and in a corridor of repeating doors that is not rare. The viewer therefore requires a second scan, with a distinct timestamp, whose alignment agrees: translations within 7\,cm and rotations satisfying $\mathrm{tr}(R_1^\top R_2)>2.996$, corresponding to a relative rotation angle below approximately $3.62^\circ$. Until both pass, the archived map stays available for inspection and the viewer withholds new writes. The rule costs time and is chosen because the failure modes are not symmetric---a delay is recoverable, whereas a misplaced scan becomes part of the record and is indistinguishable from structure actually observed. It is not sufficient in principle: scans from nearly the same place are not independent, and a statistical check cannot catch a failure that is geometric, such as furniture moved since the archive was written.

\section{Calibration experiment}\label{sec:exp}

The recorded sessions support one direct comparison, and it constitutes the quantitative evaluation reported here. Because both transforms are applied to the same returns, the measurement is paired: a change in error is attributable to the correction, not to the data.

\subsection{Data, metrics and protocol}

We reprocessed the original board returns from sessions A and B recorded on 25 November 2026. Within each session, placements 01--07 were used for fitting and 08--10 held back for selection; the deployed correction came from session A, selected on the six held-out placements pooled across both sessions. Three metrics are reported per placement: the 95th percentile of absolute point-to-plane distance, computed within a placement so that one well-observed board cannot conceal a poorly observed one; the in-plane midpoint error, a two-dimensional quantity that should not be read as a translation error of the extrinsic; and the unsigned angle between the board normal from the returns and the normal implied by the camera estimate. The limits are 30\,mm, 38\,mm and $4^\circ$, chosen so that a corrected mount does not visibly double a doorframe at the viewer's working range---engineering limits rather than sensor specifications.

The six placements reported below were also used to select the correction. Applying both transforms to the same returns gives a paired comparison on this selection set. Accuracy on new placements remains to be measured.

\subsection{Results}

\begin{figure}[t]\centering
\includegraphics[width=\linewidth]{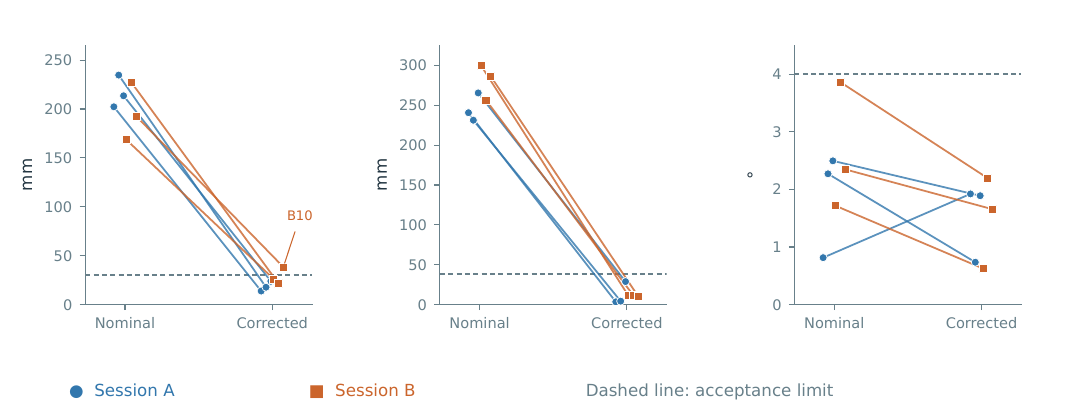}\par\vspace{2pt}
\begin{minipage}[t]{.329\linewidth}\centering\small (a) Plane-distance P95\end{minipage}\hfill\begin{minipage}[t]{.329\linewidth}\centering\small (b) In-plane centre error\end{minipage}\hfill\begin{minipage}[t]{.329\linewidth}\centering\small (c) Normal-angle error\end{minipage}
\caption{Paired errors for the six fixed-posture placements used in calibration selection. (a) Per-placement 95th percentile of absolute plane distance. (b) In-plane centre error. (c) Unsigned normal-angle error. Each line joins the nominal and corrected transforms for the same returns. Dashed lines mark the respective acceptance limits. B10 remains above the plane-distance limit in (a).}\label{fig:calresults}
\end{figure}

Table~\ref{tab:result} summarises the comparison and Fig.~\ref{fig:calresults} shows the paired errors placement by placement. Averaged over the six, the plane P95 (Fig.~\ref{fig:calresults}(a)) falls from 206.52\,mm to 23.53\,mm, a factor of 8.8; the midpoint error (Fig.~\ref{fig:calresults}(b)) falls from 263.33\,mm to 11.65\,mm; and the normal angle (Fig.~\ref{fig:calresults}(c)) falls from $2.25^\circ$ to $1.50^\circ$. Five placements meet all three limits after correction, against none before.

\begin{table}[t]\centering\small
\centerfloatcaption
\caption{Fixed-posture calibration on 25 November 2026. Error columns are arithmetic means over the six selection placements, including failures. Plane P95 is computed separately within each placement. Pass counts placements meeting all three limits.}\label{tab:result}
\begin{tabular}{@{}cccccc@{}}\toprule
Mount & Boards & Plane P95 (mm) & Center (mm) & Normal ($^\circ$) & Pass\\\midrule
Nominal & 6 & 206.52 & 263.33 & 2.25 & 0/6\\
Corrected & 6 & 23.53 & 11.65 & 1.50 & 5/6\\\bottomrule
\end{tabular}
\end{table}

The exceptions are more informative than the averages. B10 retains a plane P95 of 38.30\,mm, above the limit, while its midpoint error of 10.57\,mm is among the smallest; good in plane and poor in distance points to returns that do not belong to the board face, most plausibly from structure behind it or from the board's own edge. A08 moves the other way on orientation, its normal angle rising from $0.81^\circ$ to $1.92^\circ$ as its plane distance falls from 202.25\,mm to 13.81\,mm. The correction thus does not improve every measure at every placement, and Table~\ref{tab:result} conceals what Fig.~\ref{fig:calresults} shows. With six placements we cannot say whether B10 is an outlier or evidence of a systematic weakness in the solver.

At the reference posture, applying the fitted correction reduced the mean plane P95 and the mean in-plane midpoint error on the six selection placements. All six were recorded at the same joint configuration, so the comparison leaves the accuracy of the kinematic correction at other postures open.

\section{Map inspection on the robot: an office doorway}\label{sec:doorway}

\begin{figure}[!htbp]
\centering
\begin{minipage}[b]{.445\linewidth}\centering
\includegraphics[width=\linewidth]{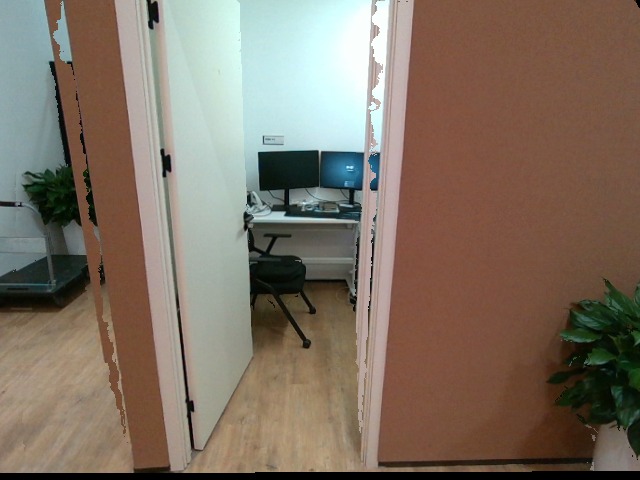}\par
\vspace{2pt}{\small (a) RGB-D colour image}\end{minipage}\hfill\begin{minipage}[b]{.534\linewidth}\centering
\includegraphics[width=\linewidth]{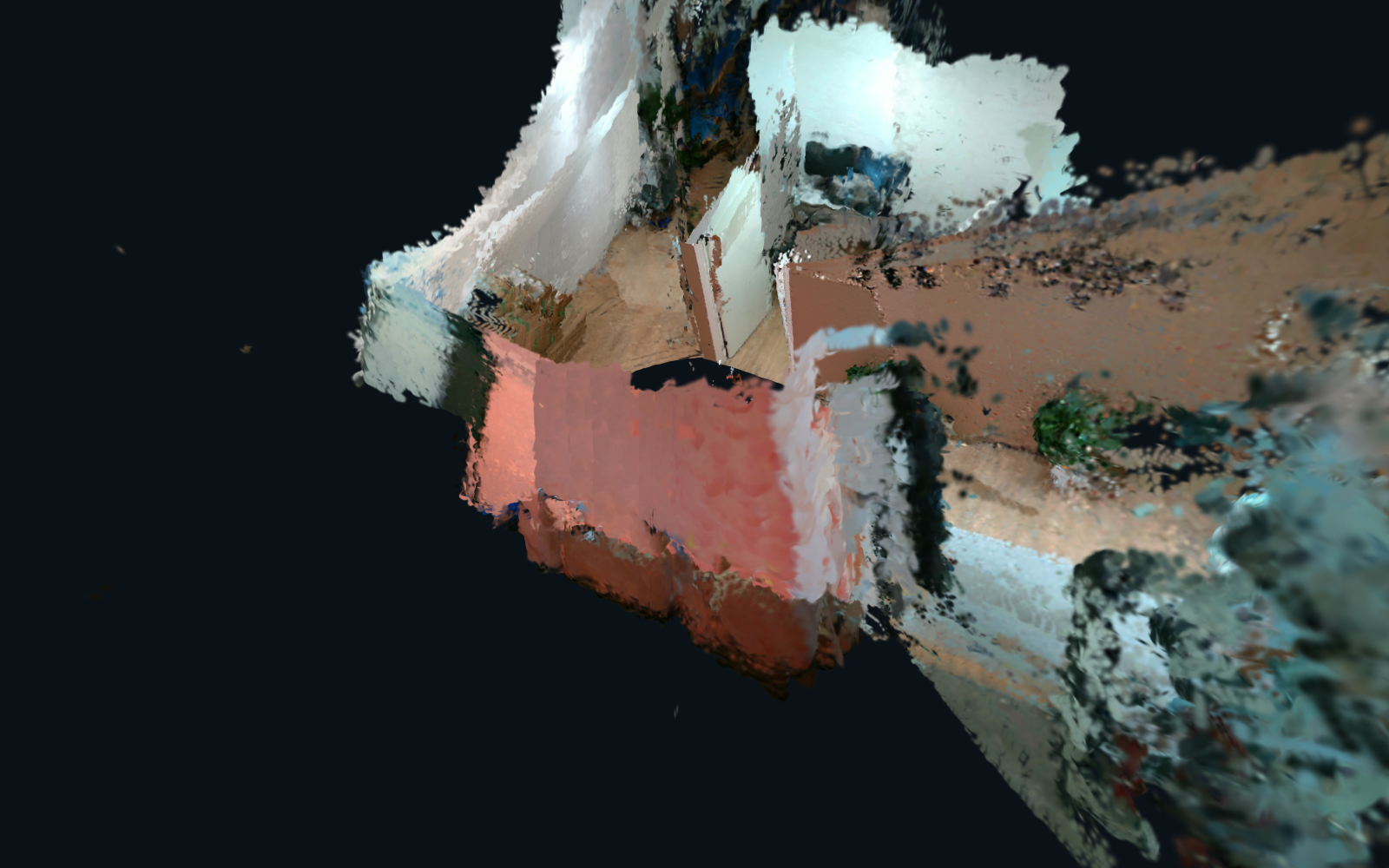}\par
\vspace{2pt}{\small (b) Doorway in the Gaussian map}\end{minipage}
\par\vspace{6pt}
\begin{minipage}[b]{.4895\linewidth}\centering
\includegraphics[width=\linewidth]{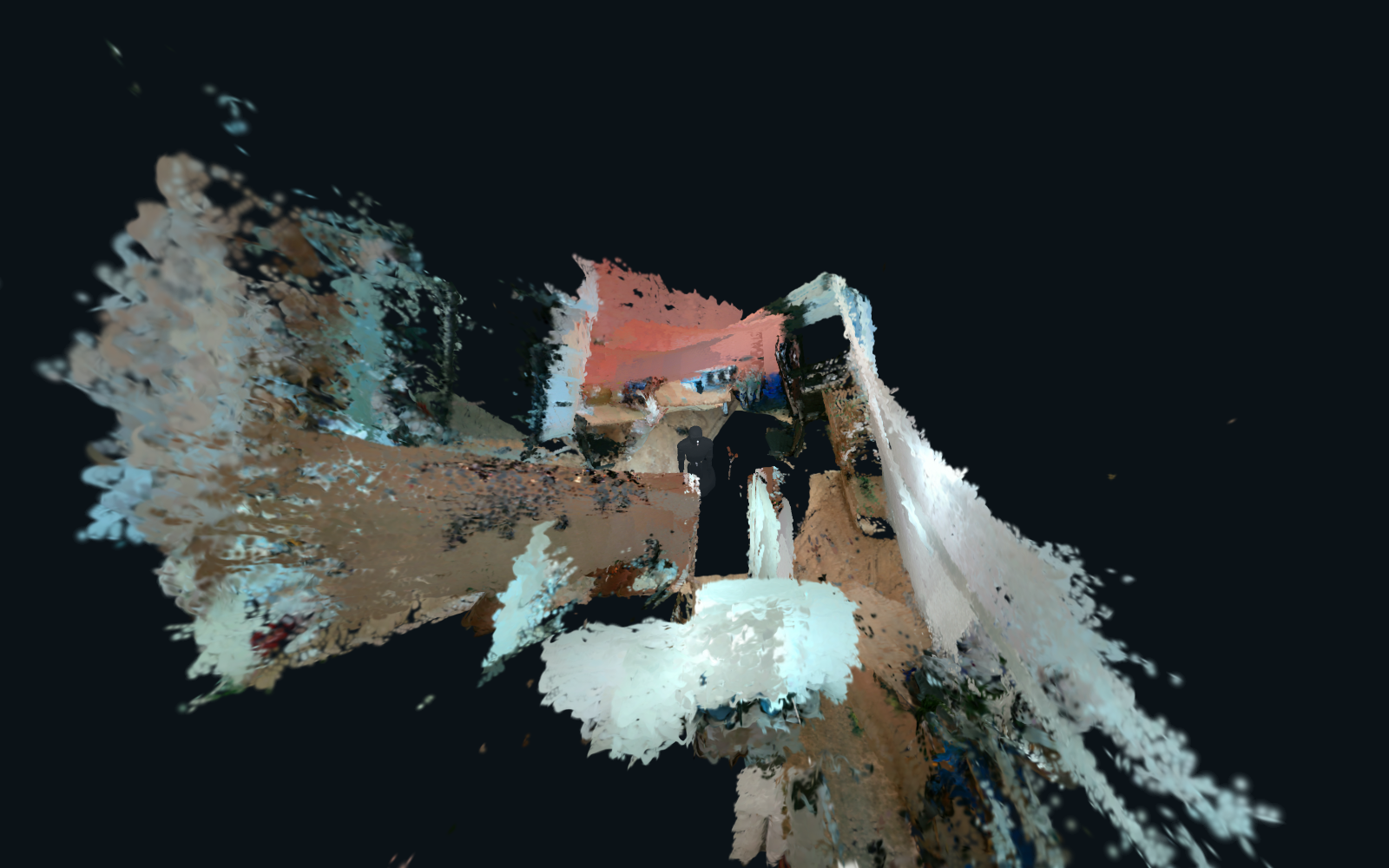}\par
\vspace{2pt}{\small (c) Gaussian map}\end{minipage}\hfill\begin{minipage}[b]{.4895\linewidth}\centering
\includegraphics[width=\linewidth]{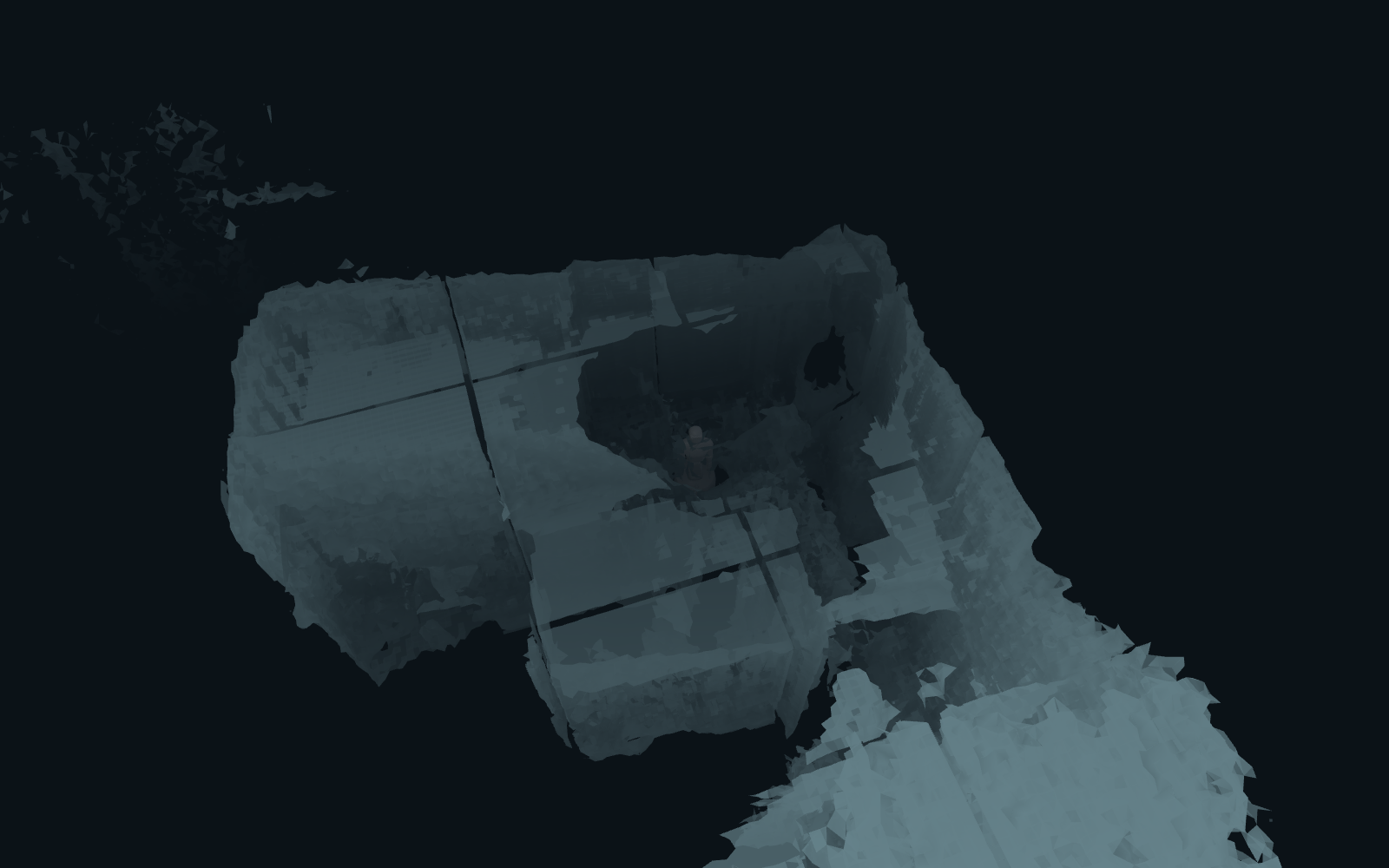}\par
\vspace{2pt}{\small (d) Accumulated LiDAR map}\end{minipage}
\caption{An office doorway recorded during one mapping session on the XR2. (a) Colour image from the head camera. (b) Oblique rendering of the doorway. (c) Gaussian appearance and (d) accumulated LiDAR geometry displayed successively with the virtual camera position, orientation and viewing distance held fixed. The panels were acquired in sequence.}\label{fig:office-doorway}
\end{figure}

Section~\ref{sec:layers} describes how the viewer stores appearance and range. On the robot, an operator first needs to relate a surface in the map to an object in the camera view. Figure~\ref{fig:office-doorway}(a) provides a reference: the head camera looks through an open door towards two monitors on a desk. From an oblique viewpoint above the doorway, the pale door, the adjoining wall and the monitors can also be identified in the Gaussian map (Fig.~\ref{fig:office-doorway}(b)). This viewpoint also reveals breaks in the reconstructed surfaces, particularly along the wall outlines.

A wider view allows the two map representations to be compared over the same area. In Fig.~\ref{fig:office-doorway}(c), wall colour helps locate the doorway within its surroundings. Switching to accumulated LiDAR returns reveals the extent and corners of the wall surfaces in Fig.~\ref{fig:office-doorway}(d). Holding the virtual viewpoint fixed allows these contours to be compared directly. The map continued to update between captures, and dark regions indicate areas where the viewer drew no surface from this viewpoint. The comparison illustrates what an operator can inspect in each layer. Measuring the geometric error at the doorframe would require an independent reference.

\section{Discussion and limitations}\label{sec:disc}

Several effects could contribute to the residual error at the reference posture, including depth and range noise and partial observations of the board edge. A fitted extrinsic can also absorb errors in the nominal kinematic chain, such as a joint offset. The fixed-posture measurements do not distinguish these effects. Measurements at other head configurations would show how well the fitted correction transfers.

Several limits follow. The calibration is a single-platform result; a second XR2 would need its own correction, since the absorbed kinematic error belongs to that machine. The estimate assumes the target was observed at a posture representative of use. Recovery searches near a seed and assumes enough shared geometry to constrain the new scan; where furniture has moved, or in a corridor of identical doors, that assumption can fail without the residual statistics showing it, and the two-scan rule reduces the risk rather than removing it. Three further limits concern the components described but not quantified: rendering suppresses sparse returns joined into surfaces, so a thin object may be absent from the view while present in the data; the fitter depends on valid depth, so reflective or textureless surfaces contribute fewer primitives than their area suggests; and storage is not atomic, so a power loss can leave the snapshot, the archive and the saved pose at different save times.

On the robot, the components beyond calibration have been exercised rather than evaluated: the four views of Table~\ref{tab:layers} have been displayed from a single archive, and a restart in place has recovered the archived frame and resumed writing. These are single runs without repeats, an external pose reference or a controlled failure, so we report them as demonstrations rather than measurements. Quantifying them requires three experiments differing mainly in what they hold fixed. A cross-posture study would repeat the paired comparison after changing head pitch, head yaw, waist angle or lift height, holding each configuration during capture and reporting the worst placement rather than the average. An archival study would replay identical poses into the native archive and into a coloured-point representation and render both from a fixed camera before and after reload, separating storage fidelity from reconstruction error. A restart study would place the robot at measured offsets from the saved pose and at a similar-looking location, comparing the saved seed alone, a single-scan rule and the two-scan rule against an external pose reference; only it can detect a confident mistake.

\section{Conclusion}

Robot View places a camera on an articulated head and a LiDAR on the chassis of a mobile robot in a common frame, keeps surface appearance in a record separate from the range measurements that support it, and requires agreement between two scans with distinct timestamps before a reopened map accepts new observations. Across the six board placements used to select the mount correction, the mean 95th percentile of absolute plane distance falls from 206.52\,mm to 23.53\,mm, and five of the six placements meet all three acceptance limits. Whether the same correction survives a change of posture, and how reliably a restarted viewer can tell a correct resumption from a confident mistake, are the questions the design was built to allow.

\section*{Use of generative AI}
Generative AI tools assisted with drafting and revising the manuscript. ImageGen assisted with the system illustration in Fig.~\ref{fig:architecture}.

\end{document}